\documentclass[sigplan,screen,nonacm]{acmart}

\begin{document}
\title{A Sui Generis QA Approach using RoBERTa for Adverse Drug Event Identification}


\author{Harshit Jain}
\authornote{Mentioned in alphabetical order of given names. All authors contributed equally to this research.}
\email{jharshit.nitj@gmail.com}

\affiliation{%
  \institution{ZS Associates}
  \city{Bengaluru}
  \state{India}
}

\author{Nishant Raj}
\authornotemark[1]
\email{nishant.raj.iitr@gmail.com}
\affiliation{%
  \institution{ZS Associates}
  \city{Bengaluru}
  \state{India}
}

\author{Suyash Mishra}
\email{smnitrkl50@gmail.com}
\authornotemark[1]
\affiliation{%
  \institution{ZS Associates}
  \city{London}
  \state{United Kingdom}
}


\setlength{\parskip}{0.06cm}
\begin{abstract}
  Extraction of adverse drug events from biomedical literature and other textual data is an important component to monitor drug-safety and this has attracted attention of many researchers in healthcare. Existing works are more pivoted around entity-relation extraction using bidirectional long short term memory networks (Bi-LSTM) which does not attain the best feature representations. In this paper, we introduce a question answering framework that exploits the robustness, masking and dynamic attention capabilities of RoBERTa by a technique of domain adaptation and attempt to overcome the aforementioned limitations. Our model outperforms the prior work by 9.53\% F1-Score.
\end{abstract}



\maketitle

\section{Introduction}
Recent advancements in drug development and approval have bolstered our healthcare ecosystem. However, this rapid paced development is accompanied by an increase in associated risks. Adverse Drug Events (ADEs) form an integral component of those risks. An ADE is defined as "an injury resulting from a medical intervention related to a drug"\cite{paper1}. 

These events create an economic burden over the system. A study by Rocchiccioli et al. showed "a statistically significant increase in all direct costs (inpatient,outpatient, therapy) during post-ADE period (+US\$1310 for all ADEs and +US\$1983 for preventable ADEs, versus the pre-event period)" \cite{paper2,paper3,paper4}. Further, national estimates are indicative of the fact that ADEs contribute at least an additional US\$ 30 billion to US healthcare costs \cite{paper5,paper6}. Thus, it becomes an utmost requirement to identify these ADEs at an early stage from biomedical literatures, EHR data and other sources in order to avoid additional costs incurred in patient management while improving pharmacovigilence practices at the same time. 

Several Adverse Drug Events have been reported to U.S. Food \& Drug Administration (FDA) through Federal Adverse Event Reporting System (FAERS). These reports submissions are voluntary and hence they may suffer from cases of massive under-reporting \cite{paper7}. Hence, researchers have started moving towards more automated approaches in machine learning. There has been a gradual shift towards using natural language processing (NLP) based methods. Early attempts have incorporated the use of resources like NLM's MetaMap, Unified Medical Language System (UMLS) etc. \cite{paper8} to extract drugs for ADE identification tasks. However, a major limitation of these approaches is that they are not able to capture the causal relationships between drug and ADE properly.

Another popular approach in the community has been to treat this problem as entity recognition and relation identification task. This approach has further been tackled using either a \textit{1. pipeline method} where entity recognition tasks are done first followed by relation identification or \textit{2. a joint training method} where training weights are shared between these two sub-tasks so that errors don't accumulate \cite{paper11,paper12,paper13}. Most of these works have utilized different variants of long short term memory (LSTM) networks for both entity recognition and relation identification tasks. Though these networks have shown promising results, they fail to capture best feature representations.

There have been ground-breaking developments in NLP with the advent of transformer architectures like BERT \cite{paper16}, GPT \cite{paper22}, RoBERTa \cite{paper18} etc. Transformer networks utilize the power of multi-head self-attention mechanism to capture context-sensitive embeddings and interactions between tokens. Since they have been pre-trained on extremely large documents for language modeling task, they have shown state of the art results with downstream tasks like sentiment analysis, question answering, machine translation etc.

Our work exploits the use of \textit{"RoBERTa"} architecture for ADE task using a question-answering(QA) framework. There are three components that lead to the final identification of ADE: \textit{NER module}, \textit{Classification module} to identify relevant texts and most important component i.e. \textit{a Query module} fine-tuned for establishing relation between drug and the ADE. There has been very limited work in this direction that leverages the transformer architecture in QA setup to extract entity-relation especially in context of ADE identification using a drug. 

Rest of our work is organized as follows. Section 2 contains related work. In Section 3, we discuss our approach in detail i.e. system architecture, dataset,  experimental setup, training and evaluation metrics. Section 4 discusses the experimental results which is followed by a conclusion in Section 5.

\section{Related Work}

Initial works in NLP focussed on the use of statistical features or kernel-based methods for relation extraction or classification related tasks \cite{paper8, paper9}. These research works focus on a pipeline approach where entities are identified first using a NER module and then a relationship is identified between those entities. Explicitly computing these statistical features poses a requirement for a large amount of annotated documents for training purposes and they are also associated with risk of errors propagating to different steps.

Li \& Ji \cite{paper10} tackled the joint entity relation extraction problem as a single transition based model using integer linear programming method to draw inferences. Deep Learning methods have recently picked up a lot of pace and contributed a lot in field of NLP. In biomedical domain, the task of NER to identify drug, disease, dosages etc. has been worked out as a sequence labelling problem where tokens are tagged according to BILOU/BIO scheme. Miwa \& Bansal \cite{paper11} proposed an end-to-end joint relation extraction model where they stacked bidirectional tree-structured LSTMs on bidirectional sequential LSTMs. Their work inspired many researchers to take this knowledge to biomedical entity and relation extraction tasks. 

Li et. al \cite{paper12} proposed a neural joint model for ADE relation extraction from biomedical texts. They utilized Bi-LSTM architecture for biomedical entity recognition using concatenation of character representation, POS embedding and word embedding as input features. This layer shared partial parameters with another Bi-LSTM layer which was tuned during a joint training process. These works using Bi-LSTMs helped to make a significant progress in entity-relation extraction tasks. However, the multi-head self attention capabilities allowed transformers to capture long range dependencies efficiently. This along with contextual dense representations from pre-training using a very large corpus allowed a better feature extraction capability.

Li et. al \cite{paper13} in their work formalize a QA framework using BERT architecture. Their research emphasizes how question based queries can encode the important information for entity-relation class identification in the question and at the same time provide a natural way of jointly modeling entity and relation. Eberts and Ulges \cite{paper14} present a span-based joint entity and relation extraction model with transformer pre-training. They add a span classification layer that filters entities from non-entities. The filtered entities are then used for relation identification purposes. We build upon the learnings of these research endeavours to devise a new QA framework for ADE identification task using a more powerful transformer architecture RoBERTa that we describe in the next section.

\section{Approach}
In this section, we introduce our system architecture (Figure 1) and explain different modules it invokes.

\begin{figure}[H]
  \centering
  \includegraphics[width=\linewidth]{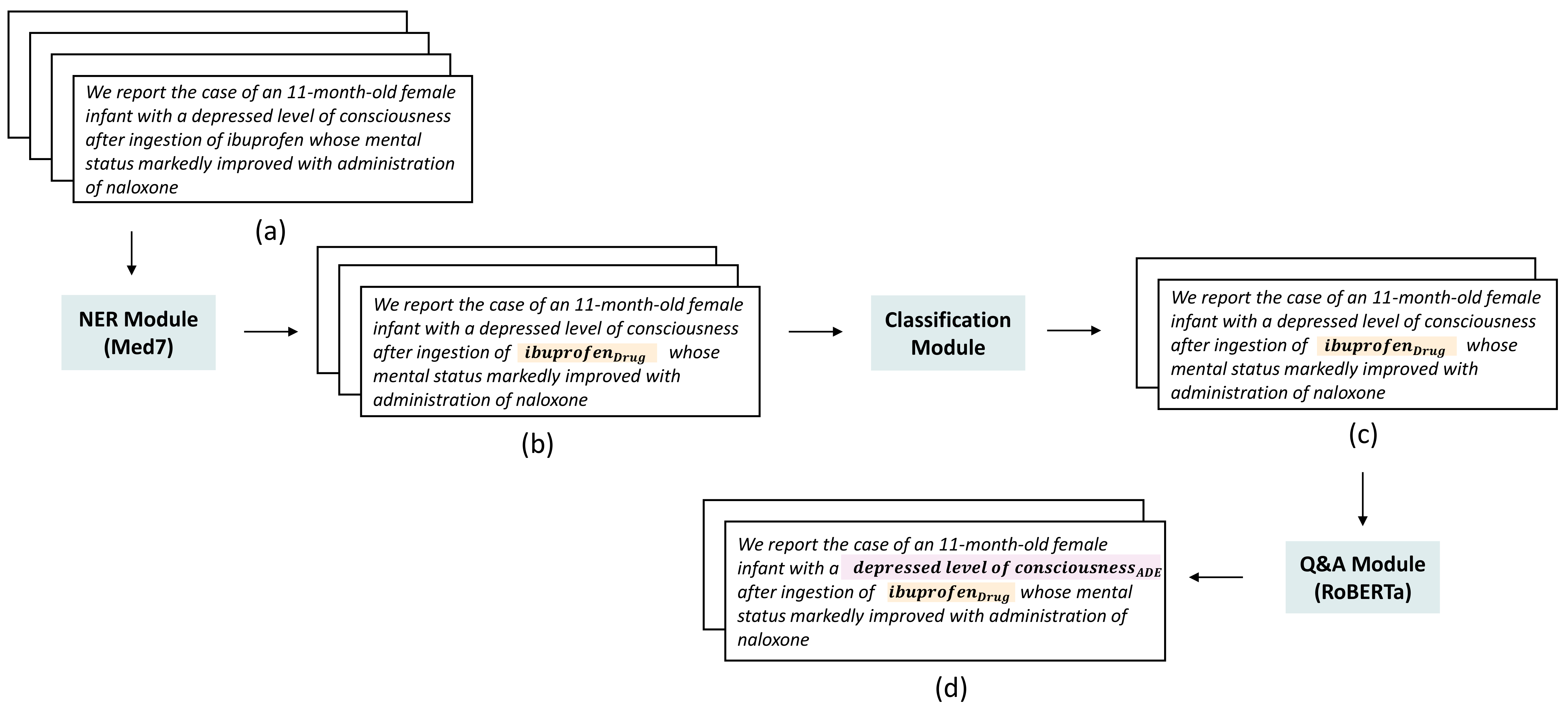}
  \caption{\textbf{End-to-End System Architecture} (a) Positive (ADE) and Negative (Non-ADE) sentences are passed through NER module (b) NER module filters sentences which don't have drug entity mentions (c) Selected sentences at NER stage are passed through Classification Module which filters out sentences where Drug-ADE relation is less probable (d) RoBERTa QA Module identifies ADE relationship \textit{(e.g. depressed level of consciousness)} corresponding to the selected drug entity \textit{(e.g. ibuprofen)}}
  
\end{figure}

\subsection{Entity Recognition Module}
Name entity recognition has been identified as a pivotal task in NLP. Classification of words from biomedical text into predefined categories like drug, disease, dosage etc. is a challenging problem. Many researchers have faced an issue of unavailability of annotated medical corpus due to which model generalization becomes difficult.

 In our system, to identify the drug names in a given phrase, we leveraged recently developed Med7 \cite{paper15} NER module which is trained on a collection of 2 million free-text patients' record from MIMIC-III corpus followed by fine-tuning on the NER task. \textit{"It has attained a lenient micro-average F1 of 0.957 across seven different categories (Dosage, Drug, Duration, Form, Frequency, Route, Strength)}".

\subsection{Classification Module}
Denoising and extraction of relevant information in textual data is a crucial step as it improves the learning mechanism and generalizing capability of any model. After recognizing the drug entities (section 3.1), to identify the phrases where at least one drug and adverse event pair coexists at stage 2, we trained a Bi-LSTM \cite{paper23} based binary classifier on ADE sentences (section 4.1) and cross-validated it in K-Fold setting.

This model aims to filter out phrases with no presence of drug and adverse event pair and helps in improving the performance of our Q\&A module and thereby making it more robust.

\begin{figure*}[h]
  \centering
  \includegraphics[width=\linewidth]{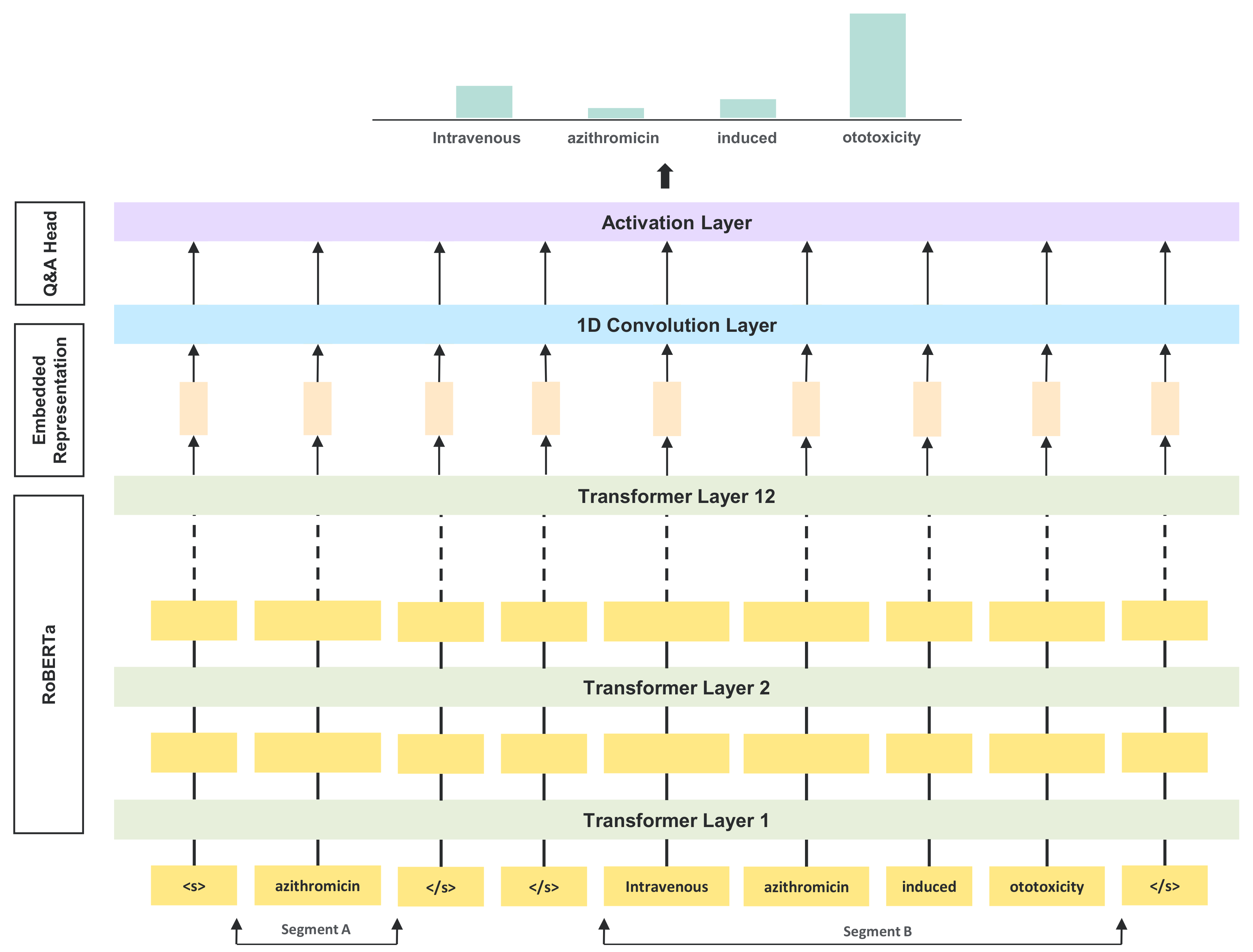}
  \caption{\textbf{Q\&A Module for Adverse Drug Event Identification}. Input segment A consists of a drug acting as an alias for question and segment B consists of context where an ADE is mentioned. Encoded representation of segments passed as an input through 12-layered transformer network. 1-D convolutional layer is applied on top of 768 dimensional embedded representations followed by a softmax activation to identify the ADE.}
  
\end{figure*}

\subsection{Q\&A Module}
Given a passage of text with a user query, a question-answering system discovers the span of text in the passage that best describes answer to the question being asked.

BERT \cite{paper16} based pre-trained language models have achieved state-of-the-art performances on multiple tasks like semantic role labeling, question answering, machine translation, etc as it has been trained on large scale corpora and generalizes the downstream task very well. It can learn the extremely complex representation in text with a transformer-based \cite{paper17} self-attention mechanism and play a crucial role in improving varieties of NLP systems. 

Liu et. al \cite{paper18} in their paper that introduces RoBERTa, observed BERT to be "\textit{significantly undertrained}" and also found out that with a better selection of hyperparamerters and training size, its performance could be considerably improved.  RoBERTa, a re-implemented version of BERT in FAIRSEQ \cite{paper19}, has been trained with different learning rate, number of warm-up steps, batch size and out-performs state-of-the-art results on GLUE, RACE and SQuAD dataset. 
We use a pre-trained RoBERTa base model and fine-tune it on drug-related adverse effects corpus (section 4.1)  to identify the adverse event corresponding to a drug by adding a Q\&A head (Figure 2). First we process the data in desired input format for RoBERTa into 2 segments A and B. Segment A consists an encoded vector of drug treated as a question followed by segment B that consists another encoded vector of context/sentence where adverse event is mentioned. Then we pass this processed data into a 12-layered transformer network of RoBERTa and use its output that represents the 768 dimensional learnt embeddings of encoded input for further processing. After this, we apply a one dimensional CNN layer with a (1 x 1) convolution filter that creates a feature map of these embeddings followed by a softmax activation layer to predict the probability of start/end tokens of the adverse event present in a span of the given text.

\section{Experimental Studies}
\subsection{Dataset and Evaluation Metrics}
We use drug-related adverse effects corpus \cite{paper20} containing sentences from 1644 PubMed abstracts. These sentences are divided into 2 categories (i) ADE (ii) Non-ADE. Former consists of sentences where at least one pair of drug and its adverse effect is present while latter consists of sentences with no such pair. 
\begin{table}[H]
  \caption{Distribution of sentences in dataset}
  \label{tab:freq}
  \begin{tabular}{ccl}
    \toprule
    Category& \ Number of Unique Sentences\\
    \midrule
    \ ADE & 6617\\
    \ Non-ADE & 16688\\
  \bottomrule
\end{tabular}
\end{table}

Examples for ADE and Non-ADE instances are:
\begin{itemize}
    \item \textbf{ADE:} \textit{14-year-old girl with newly diagnosed sle developed a \textbf{pruritic bullous eruption} while on \textbf{prednisone}}
    \setlength{\itemsep}{0.2pt}
     \item \textbf{Non-ADE:} \textit{This patient did not have any predisposing factors for the development of an aortic thrombus before the chemotherapy was initiated.}
\end{itemize}
To gauge the performance of our system, we use common performance metrics such as Precision, Recall and F1. 

\begin{equation}
P = \frac{TP}{TP+FP}, \\
R = \frac{TP}{TP+FN}, \\
F1 = \frac{2*P*R}{P+R}
\end{equation} 

\subsection{Training}
We perform training for 2 different modules involved in our system i.e. Classification and Q\&A. 

(i) For classification module, we construct a train and test dataset by selecting randomly sampled 9931 training and 1272 testing instances. We train a Bi-LSTM with Adam optimizer \cite{paper21} minimizing binary-crossentropy loss in stratified K-Fold (k=10) setting to ensure consistency in our model performance. For final prediction on hold out dataset, we use an ensemble of these 10 Bi-LSTMs. Below appears the distribution of target variable in train and test sets:

\begin{table}[H]
  \caption{\centering{Distribution of target variable in train and test for Classification Module}}
  \label{tab:freq}
  \begin{tabular}{ccl}
    \toprule
    Positive (ADE) &\ Negative (Non-ADE ) &\ Dataset \\
    \midrule
   3976 &\ 5955 &\ Train \\
   610 & \ 662 &\ Test \\
  \bottomrule
\end{tabular}
\end{table}

(ii) To predict ADE corresponding to a drug, we create the train \& and test data by determining 5955 training and 662 testing instances from ADE sentences (section 4.1) with random sampling. 

We use pre-trained RoBERTa base model and finetune it on these 5955 sentences with a 1D CNN head (Figure 2) followed by a softmax activation layer to generate probability corresponding to start and end token of adverse event present in a context. 

We train this entire architecture in K-Fold setting (k=5) with 3 epochs each on 12GB Nvidia P100 GPU.
We use an ensemble of prediction probabilities for start \& end tokens generated by models trained on each of the 5 folds. We set a learning rate of 3e-5 for Adam optimizer and categorical-crossentropy as our loss function with a label smoothing of 0.1. 

\section{Results}
Our RoBERTa based QA approach utilizes \textit{drug} entity passed as a \textit{question} to determine \textit{answer} i.e. \textit{adverse drug event} in the given context. We determine the performance at two levels: (i) Performance of individual modules (ii) Performance of entire architecture (system). 

 Previous works \cite{paper12, paper13, paper14} in the field focus on approaches that leverage either a joint training approach or a cascading pipeline approach to identify entities and then classify those extracted entities to ascertain existence of any relationship. However, our approach as described in previous section is not tailored similar to these settings. Hence, a direct comparison of individual modules is difficult. We elucidate our approach for calculation of system's performance using these three modules together. 
 
\subsection{Performance of Individual Modules} Table 3 below details the performance of Classification and QA module corresponding to their selected training and validation set as described in the previous section. The results for classification module are reported based on mean results from 10-Fold validation sets. The presence of Non-ADE sentences (Section 4.1) where drug is either not known or wrongly identified is expected to create a bias in understanding of effectiveness of RoBERTa QA module for ADE identification tasks. Hence, we use only ADE sentences where drug is known beforehand for determining the performance of Roberta QA framework. In this scenario, the precision would equal to 1 and hence we use recall to gauge the true efficacy of QA module.

\begin{table}[H]
    \caption{\centering{Performance Evaluation of Classification \& QA Modules}}
    \label{tab:freq}
    \begin{tabular}{ccl}
      \toprule
        {Metrics} &\ {Classification Module} &\ {QA Module} \\
      \midrule
          {Precision} &\ 82.74  &\ - \\
          {Recall} &\ 81.44 &\ 87.37 \\
          {F1} &\ 82.06 &\ - \\
     \bottomrule
    \end{tabular}
\end{table}

\subsection{Performance of End-to-End Architecture}  Errors generated by different components in a system create a cascading effect and this aggregation of errors might render the system to a practically futile state. 

In real-world applications, the overall task comprises of drug identification, noise removal and then drug-ADE relationship identification. In that process, obliteration of noisy textual information should also be accounted into the success criteria for smooth functioning of a NLP pipeline. We describe our approach to calculate the efficacy of entire system through Figure 3 below. 

\begin{figure}[H]
  \centering
  \includegraphics[width=\linewidth]{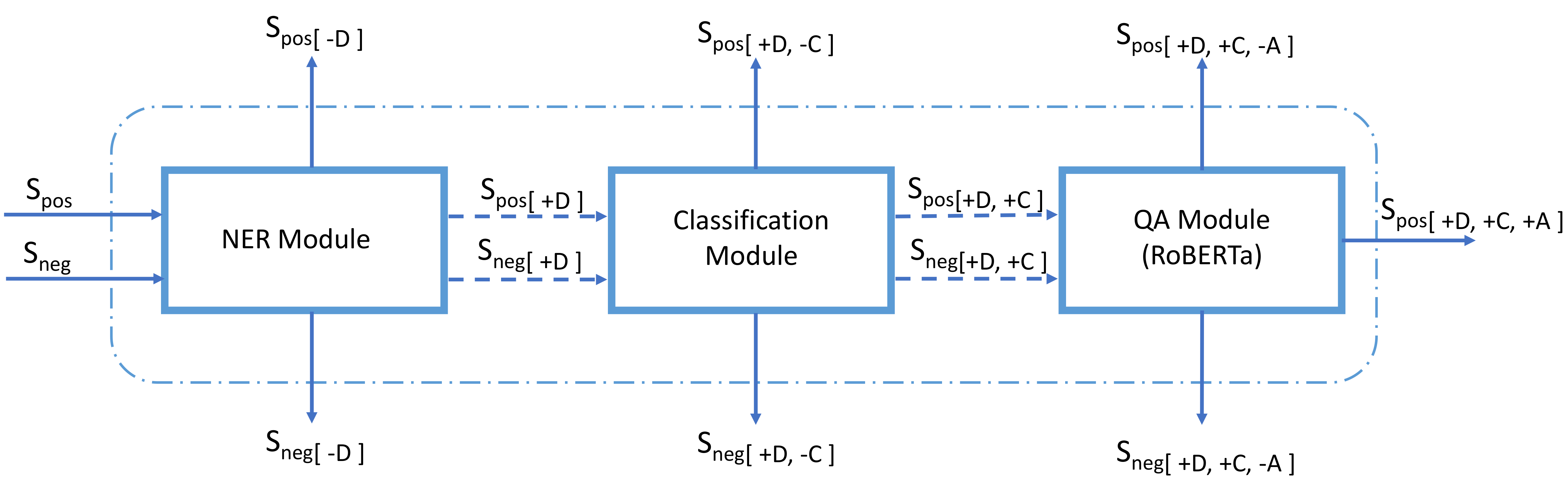}
  \caption{\textbf{End-to-End System Architecture's Overall Performance Calculation.} Sentences in two sets (S\textsubscript{pos} \& S\textsubscript{neg}) are passed into the system with an objective to identify drug-ADE relationship from S\textsubscript{pos} and removal of all the S\textsubscript{neg} instances. Classification matrix generated at each stage is leveraged for final calculation of precision, recall and F1 metrics.}
\end{figure}

At each stage either a sentence is eliminated from system or it moves to next module for operation. S\textsubscript{pos} [+D,+C,+A] denotes that among the positive sentences passed as input, these sentences had drug and were sent to QA module by classification module where ADE was accurately identified by RoBERTa QA architecture. S\textsubscript{neg}[-D], S\textsubscript{neg}[+D,-C] contribute to correctly removed instances from system. Similar analogy for their S\textsubscript{pos} counterparts identifies misclassified samples from the system. Equations for calculation of Precision, Recall and F1 scores for the entire architecture can be calculated using (1) and the modified definitions below:

\begin{gather}
TP = S_{neg[ -D ]}+S_{neg[ +D, -C ]}+S_{pos[ +D, +C, +A ]} \nonumber\\
FN = S_{pos[ -D ]}+S_{pos[ +D, -C ]}\\
FP = S_{pos[ +D, +C, -A ]}+S_{neg[ +D, +C, -A ]}\nonumber
\end{gather}

We detail the final performance metrics for the end-to-end system architecture in Table 4. We also visit the effectiveness of prominent approaches for ADE identification task in Table 5 below. After a thorough study of the relevant literature for ADE identification tasks, we observed that in joint training approaches, reporting of results is done in a way that the relation classification task also incorporates effects due to errors generated by entity recognition modules. Hence, we compare the results of relation classification task to the effectiveness of our entire end-to-end architecture.

\begin{table}[H]
    \caption{\centering{End-to-End System Performance }}
    \label{tab:freq}
    \begin{tabular}{ccl}
      \toprule
         {P} &\ {R} &\ {F1} \\
      \midrule
          {88.37} &\ 84.44  &\ 86.36 \\
     \bottomrule
    \end{tabular}
\end{table}

\begin{table}[H]
    \caption{Comparison with different methods}
    \label{tab:freq}
    \begin{tabular}{cccc}
         \toprule
         {Methods} &\ {P} &\ {R} &\ {F1} \\
         \midrule
         {CNN + Global features \cite{paper24}} &\ {64.00} &\ {62.90} &\ {63.40} \\
         {BiLSTM + SDP \cite{paper12}} &\ {67.50} &\ {75.80} &\ {71.40} \\
         {SpERT \cite{paper14}} &\ {77.77} &\ {79.96} &\ {78.84} \\
         \textbf{Our Model (End-to-End)} &\ \textbf{88.37} &\ \textbf{84.44} &\ \textbf{86.36} \\
         \bottomrule
    \end{tabular}
\end{table}

With a closer and detailed inspection, we observe that even after error propagation effect, the overall effectiveness of our system is better than existing approaches for ADE identification.

\section {Conclusion}
In this paper, we propose a novel approach for ADE identification tasks in biomedical texts. The experimental results highlight the effectiveness of using an end-to-end pipeline comprising of NER, classification and RoBERTa based QA modules where it achieves competitive performances in comparison to the existing best systems. We hope that our work would open up a new dimension in ADE identification and entity-relation tasks in general.

Although our end-to-end architecture achieves promising results, we plan to improve on the NER component in the architecture by building a transformer based biomedical NER. We also plan to extend this pipeline to social media platforms like twitter for pharmacovigilance based studies.

\bibliographystyle{ACM-Reference-Format}
\bibliography{sample-base}


\begin{thebibliography}{24}


\ifx \showCODEN    \undefined \def \showCODEN     #1{\unskip}     \fi
\ifx \showDOI      \undefined \def \showDOI       #1{#1}\fi
\ifx \showISBNx    \undefined \def \showISBNx     #1{\unskip}     \fi
\ifx \showISBNxiii \undefined \def \showISBNxiii  #1{\unskip}     \fi
\ifx \showISSN     \undefined \def \showISSN      #1{\unskip}     \fi
\ifx \showLCCN     \undefined \def \showLCCN      #1{\unskip}     \fi
\ifx \shownote     \undefined \def \shownote      #1{#1}          \fi
\ifx \showarticletitle \undefined \def \showarticletitle #1{#1}   \fi
\ifx \showURL      \undefined \def \showURL       {\relax}        \fi
\providecommand\bibfield[2]{#2}
\providecommand\bibinfo[2]{#2}
\providecommand\natexlab[1]{#1}
\providecommand\showeprint[2][]{arXiv:#2}

\bibitem[\protect\citeauthoryear{Bates, Cullen, Laird, Petersen, Small, Servi,
  Laffel, Sweitzer, Shea, Hallisey, Vander~Vliet, Nemeskal, Leape, Bates,
  Hojnowski-Diaz, Petrycki, Cotugno, Patterson, Hickey, Kleefield, Cooper,
  Kinneally, Demonaco, Clapp, Gallivan, Ives, Porter, Thompson, Hackman, and
  Edmondson}{Bates et~al\mbox{.}}{1995}]%
        {paper1}
\bibfield{author}{\bibinfo{person}{David~W. Bates}, \bibinfo{person}{David~J.
  Cullen}, \bibinfo{person}{Nan Laird}, \bibinfo{person}{Laura~A. Petersen},
  \bibinfo{person}{Stephen~D. Small}, \bibinfo{person}{Deborah Servi},
  \bibinfo{person}{Glenn Laffel}, \bibinfo{person}{Bobbie~J. Sweitzer},
  \bibinfo{person}{Brian~F. Shea}, \bibinfo{person}{Robert Hallisey},
  \bibinfo{person}{Martha Vander~Vliet}, \bibinfo{person}{Roberta Nemeskal},
  \bibinfo{person}{Lucian~L. Leape}, \bibinfo{person}{David Bates},
  \bibinfo{person}{Patricia Hojnowski-Diaz}, \bibinfo{person}{Stephen
  Petrycki}, \bibinfo{person}{Michael Cotugno}, \bibinfo{person}{Heather
  Patterson}, \bibinfo{person}{Mairead Hickey}, \bibinfo{person}{Sharon
  Kleefield}, \bibinfo{person}{Jeffrey Cooper}, \bibinfo{person}{Ellen
  Kinneally}, \bibinfo{person}{Harold~J. Demonaco},
  \bibinfo{person}{Margaret~Dempsey Clapp}, \bibinfo{person}{Theresa Gallivan},
  \bibinfo{person}{Jeanette Ives}, \bibinfo{person}{Kathy Porter},
  \bibinfo{person}{B.~Taylor Thompson}, \bibinfo{person}{J.~Richard Hackman},
  {and} \bibinfo{person}{Amy Edmondson}.} \bibinfo{year}{1995}\natexlab{}.
\newblock \showarticletitle{{Incidence of Adverse Drug Events and Potential
  Adverse Drug Events: Implications for Prevention}}.
\newblock \bibinfo{journal}{\emph{JAMA}} \bibinfo{volume}{274},
  \bibinfo{number}{1} (\bibinfo{date}{07} \bibinfo{year}{1995}),
  \bibinfo{pages}{29--34}.
\newblock
\showISSN{0098-7484}
\urldef\tempurl%
\url{https://doi.org/10.1001/jama.1995.03530010043033}
\showDOI{\tempurl}


\bibitem[\protect\citeauthoryear{Chan and Roth}{Chan and Roth}{2011}]%
        {paper9}
\bibfield{author}{\bibinfo{person}{Yee~Seng Chan} {and} \bibinfo{person}{Dan
  Roth}.} \bibinfo{year}{2011}\natexlab{}.
\newblock \showarticletitle{Exploiting syntactico-semantic structures for
  relation extraction}. In \bibinfo{booktitle}{\emph{Proceedings of the 49th
  Annual Meeting of the Association for Computational Linguistics: Human
  Language Technologies}}. \bibinfo{pages}{551--560}.
\newblock


\bibitem[\protect\citeauthoryear{Chiatti, Bustacchini, Furneri, Mantovani,
  Cristiani, Misuraca, and Lattanzio}{Chiatti et~al\mbox{.}}{2012}]%
        {paper4}
\bibfield{author}{\bibinfo{person}{Carlos Chiatti}, \bibinfo{person}{Silvia
  Bustacchini}, \bibinfo{person}{Gianluca Furneri}, \bibinfo{person}{Lorenzo
  Mantovani}, \bibinfo{person}{Marco Cristiani}, \bibinfo{person}{Clementina
  Misuraca}, {and} \bibinfo{person}{Fabrizia Lattanzio}.}
  \bibinfo{year}{2012}\natexlab{}.
\newblock \showarticletitle{The economic burden of inappropriate drug
  prescribing, lack of adherence and compliance, adverse drug events in older
  people}.
\newblock \bibinfo{journal}{\emph{Drug safety}} \bibinfo{volume}{35},
  \bibinfo{number}{1} (\bibinfo{year}{2012}), \bibinfo{pages}{73--87}.
\newblock


\bibitem[\protect\citeauthoryear{Devlin, Chang, Lee, and Toutanova}{Devlin
  et~al\mbox{.}}{2018}]%
        {paper16}
\bibfield{author}{\bibinfo{person}{Jacob Devlin}, \bibinfo{person}{Ming-Wei
  Chang}, \bibinfo{person}{Kenton Lee}, {and} \bibinfo{person}{Kristina
  Toutanova}.} \bibinfo{year}{2018}\natexlab{}.
\newblock \showarticletitle{Bert: Pre-training of deep bidirectional
  transformers for language understanding}.
\newblock \bibinfo{journal}{\emph{arXiv preprint arXiv:1810.04805}}
  (\bibinfo{year}{2018}).
\newblock


\bibitem[\protect\citeauthoryear{Eberts and Ulges}{Eberts and Ulges}{2019}]%
        {paper14}
\bibfield{author}{\bibinfo{person}{Markus Eberts} {and} \bibinfo{person}{Adrian
  Ulges}.} \bibinfo{year}{2019}\natexlab{}.
\newblock \showarticletitle{Span-based Joint Entity and Relation Extraction
  with Transformer Pre-training}.
\newblock \bibinfo{journal}{\emph{arXiv preprint arXiv:1909.07755}}
  (\bibinfo{year}{2019}).
\newblock


\bibitem[\protect\citeauthoryear{Edlavitch}{Edlavitch}{1988}]%
        {paper7}
\bibfield{author}{\bibinfo{person}{Stanley~A Edlavitch}.}
  \bibinfo{year}{1988}\natexlab{}.
\newblock \showarticletitle{Adverse drug event reporting: improving the low US
  reporting rates}.
\newblock \bibinfo{journal}{\emph{Archives of internal medicine}}
  \bibinfo{volume}{148}, \bibinfo{number}{7} (\bibinfo{year}{1988}),
  \bibinfo{pages}{1499--1503}.
\newblock


\bibitem[\protect\citeauthoryear{Gurulingappa, Rajput, Roberts, Fluck,
  Hofmann-Apitius, and Toldo}{Gurulingappa et~al\mbox{.}}{2012}]%
        {paper20}
\bibfield{author}{\bibinfo{person}{Harsha Gurulingappa},
  \bibinfo{person}{Abdul~Mateen Rajput}, \bibinfo{person}{Angus Roberts},
  \bibinfo{person}{Juliane Fluck}, \bibinfo{person}{Martin Hofmann-Apitius},
  {and} \bibinfo{person}{Luca Toldo}.} \bibinfo{year}{2012}\natexlab{}.
\newblock \showarticletitle{Development of a benchmark corpus to support the
  automatic extraction of drug-related adverse effects from medical case
  reports}.
\newblock \bibinfo{journal}{\emph{Journal of biomedical informatics}}
  \bibinfo{volume}{45}, \bibinfo{number}{5} (\bibinfo{year}{2012}),
  \bibinfo{pages}{885--892}.
\newblock


\bibitem[\protect\citeauthoryear{Jagannatha, Liu, Liu, and Yu}{Jagannatha
  et~al\mbox{.}}{2019}]%
        {paper5}
\bibfield{author}{\bibinfo{person}{Abhyuday Jagannatha},
  \bibinfo{person}{Feifan Liu}, \bibinfo{person}{Weisong Liu}, {and}
  \bibinfo{person}{Hong Yu}.} \bibinfo{year}{2019}\natexlab{}.
\newblock \showarticletitle{Overview of the first natural language processing
  challenge for extracting medication, indication, and adverse drug events from
  electronic health record notes (MADE 1.0)}.
\newblock \bibinfo{journal}{\emph{Drug safety}} \bibinfo{volume}{42},
  \bibinfo{number}{1} (\bibinfo{year}{2019}), \bibinfo{pages}{99--111}.
\newblock


\bibitem[\protect\citeauthoryear{Johnson and Booman}{Johnson and
  Booman}{1996}]%
        {paper6}
\bibfield{author}{\bibinfo{person}{Jeffery Johnson} {and} \bibinfo{person}{Lyle
  Booman}.} \bibinfo{year}{1996}\natexlab{}.
\newblock \showarticletitle{Drug-related morbidity and mortality}.
\newblock \bibinfo{journal}{\emph{Journal of Managed Care Pharmacy}}
  \bibinfo{volume}{2}, \bibinfo{number}{1} (\bibinfo{year}{1996}),
  \bibinfo{pages}{39--47}.
\newblock


\bibitem[\protect\citeauthoryear{Kingma and Ba}{Kingma and Ba}{2014}]%
        {paper21}
\bibfield{author}{\bibinfo{person}{Diederik~P Kingma} {and}
  \bibinfo{person}{Jimmy Ba}.} \bibinfo{year}{2014}\natexlab{}.
\newblock \showarticletitle{Adam: A method for stochastic optimization}.
\newblock \bibinfo{journal}{\emph{arXiv preprint arXiv:1412.6980}}
  (\bibinfo{year}{2014}).
\newblock


\bibitem[\protect\citeauthoryear{Kormilitzin, Vaci, Liu, and
  Nevado-Holgado}{Kormilitzin et~al\mbox{.}}{2020}]%
        {paper15}
\bibfield{author}{\bibinfo{person}{Andrey Kormilitzin},
  \bibinfo{person}{Nemanja Vaci}, \bibinfo{person}{Qiang Liu}, {and}
  \bibinfo{person}{Alejo Nevado-Holgado}.} \bibinfo{year}{2020}\natexlab{}.
\newblock \showarticletitle{Med7: a transferable clinical natural language
  processing model for electronic health records}.
\newblock \bibinfo{journal}{\emph{arXiv preprint arXiv:2003.01271}}
  (\bibinfo{year}{2020}).
\newblock


\bibitem[\protect\citeauthoryear{Li, Zhang, Fu, and Ji}{Li
  et~al\mbox{.}}{2017}]%
        {paper12}
\bibfield{author}{\bibinfo{person}{Fei Li}, \bibinfo{person}{Meishan Zhang},
  \bibinfo{person}{Guohong Fu}, {and} \bibinfo{person}{Donghong Ji}.}
  \bibinfo{year}{2017}\natexlab{}.
\newblock \showarticletitle{A neural joint model for entity and relation
  extraction from biomedical text}.
\newblock \bibinfo{journal}{\emph{BMC bioinformatics}} \bibinfo{volume}{18},
  \bibinfo{number}{1} (\bibinfo{year}{2017}), \bibinfo{pages}{1--11}.
\newblock


\bibitem[\protect\citeauthoryear{Li, Zhang, Zhang, and Ji}{Li
  et~al\mbox{.}}{2016}]%
        {paper24}
\bibfield{author}{\bibinfo{person}{Fei Li}, \bibinfo{person}{Yue Zhang},
  \bibinfo{person}{Meishan Zhang}, {and} \bibinfo{person}{Donghong Ji}.}
  \bibinfo{year}{2016}\natexlab{}.
\newblock \showarticletitle{Joint Models for Extracting Adverse Drug Events
  from Biomedical Text.}. In \bibinfo{booktitle}{\emph{IJCAI}},
  Vol.~\bibinfo{volume}{2016}. \bibinfo{pages}{2838--2844}.
\newblock


\bibitem[\protect\citeauthoryear{Li and Ji}{Li and Ji}{2014}]%
        {paper10}
\bibfield{author}{\bibinfo{person}{Qi Li} {and} \bibinfo{person}{Heng Ji}.}
  \bibinfo{year}{2014}\natexlab{}.
\newblock \showarticletitle{Incremental joint extraction of entity mentions and
  relations}. In \bibinfo{booktitle}{\emph{Proceedings of the 52nd Annual
  Meeting of the Association for Computational Linguistics (Volume 1: Long
  Papers)}}. \bibinfo{pages}{402--412}.
\newblock


\bibitem[\protect\citeauthoryear{Li, Yin, Sun, Li, Yuan, Chai, Zhou, and Li}{Li
  et~al\mbox{.}}{2019}]%
        {paper13}
\bibfield{author}{\bibinfo{person}{Xiaoya Li}, \bibinfo{person}{Fan Yin},
  \bibinfo{person}{Zijun Sun}, \bibinfo{person}{Xiayu Li},
  \bibinfo{person}{Arianna Yuan}, \bibinfo{person}{Duo Chai},
  \bibinfo{person}{Mingxin Zhou}, {and} \bibinfo{person}{Jiwei Li}.}
  \bibinfo{year}{2019}\natexlab{}.
\newblock \showarticletitle{Entity-relation extraction as multi-turn question
  answering}.
\newblock \bibinfo{journal}{\emph{arXiv preprint arXiv:1905.05529}}
  (\bibinfo{year}{2019}).
\newblock


\bibitem[\protect\citeauthoryear{Liu, Ott, Goyal, Du, Joshi, Chen, Levy, Lewis,
  Zettlemoyer, and Stoyanov}{Liu et~al\mbox{.}}{2019}]%
        {paper18}
\bibfield{author}{\bibinfo{person}{Yinhan Liu}, \bibinfo{person}{Myle Ott},
  \bibinfo{person}{Naman Goyal}, \bibinfo{person}{Jingfei Du},
  \bibinfo{person}{Mandar Joshi}, \bibinfo{person}{Danqi Chen},
  \bibinfo{person}{Omer Levy}, \bibinfo{person}{Mike Lewis},
  \bibinfo{person}{Luke Zettlemoyer}, {and} \bibinfo{person}{Veselin
  Stoyanov}.} \bibinfo{year}{2019}\natexlab{}.
\newblock \showarticletitle{Roberta: A robustly optimized bert pretraining
  approach}.
\newblock \bibinfo{journal}{\emph{arXiv preprint arXiv:1907.11692}}
  (\bibinfo{year}{2019}).
\newblock


\bibitem[\protect\citeauthoryear{Miwa and Bansal}{Miwa and Bansal}{2016}]%
        {paper11}
\bibfield{author}{\bibinfo{person}{Makoto Miwa} {and} \bibinfo{person}{Mohit
  Bansal}.} \bibinfo{year}{2016}\natexlab{}.
\newblock \showarticletitle{End-to-end relation extraction using lstms on
  sequences and tree structures}.
\newblock \bibinfo{journal}{\emph{arXiv preprint arXiv:1601.00770}}
  (\bibinfo{year}{2016}).
\newblock


\bibitem[\protect\citeauthoryear{Ott, Edunov, Baevski, Fan, Gross, Ng,
  Grangier, and Auli}{Ott et~al\mbox{.}}{2019}]%
        {paper19}
\bibfield{author}{\bibinfo{person}{Myle Ott}, \bibinfo{person}{Sergey Edunov},
  \bibinfo{person}{Alexei Baevski}, \bibinfo{person}{Angela Fan},
  \bibinfo{person}{Sam Gross}, \bibinfo{person}{Nathan Ng},
  \bibinfo{person}{David Grangier}, {and} \bibinfo{person}{Michael Auli}.}
  \bibinfo{year}{2019}\natexlab{}.
\newblock \showarticletitle{fairseq: A fast, extensible toolkit for sequence
  modeling}.
\newblock \bibinfo{journal}{\emph{arXiv preprint arXiv:1904.01038}}
  (\bibinfo{year}{2019}).
\newblock


\bibitem[\protect\citeauthoryear{Radford, Wu, Child, Luan, Amodei, and
  Sutskever}{Radford et~al\mbox{.}}{2019}]%
        {paper22}
\bibfield{author}{\bibinfo{person}{Alec Radford}, \bibinfo{person}{Jeffrey Wu},
  \bibinfo{person}{Rewon Child}, \bibinfo{person}{David Luan},
  \bibinfo{person}{Dario Amodei}, {and} \bibinfo{person}{Ilya Sutskever}.}
  \bibinfo{year}{2019}\natexlab{}.
\newblock \showarticletitle{Language models are unsupervised multitask
  learners}.
\newblock \bibinfo{journal}{\emph{OpenAI Blog}} \bibinfo{volume}{1},
  \bibinfo{number}{8} (\bibinfo{year}{2019}), \bibinfo{pages}{9}.
\newblock


\bibitem[\protect\citeauthoryear{Rocchiccioli, Sanford, and
  Caplinger}{Rocchiccioli et~al\mbox{.}}{2007}]%
        {paper2}
\bibfield{author}{\bibinfo{person}{Judith~T Rocchiccioli},
  \bibinfo{person}{Julie Sanford}, {and} \bibinfo{person}{Bonnie Caplinger}.}
  \bibinfo{year}{2007}\natexlab{}.
\newblock \showarticletitle{Polymedicine and aging. Enhancing older adult care
  through advanced practitioners. GNPs and elder care pharmacists can help
  provide optimal pharmaceutical care}.
\newblock \bibinfo{journal}{\emph{Journal of gerontological nursing}}
  \bibinfo{volume}{33}, \bibinfo{number}{7} (\bibinfo{date}{July}
  \bibinfo{year}{2007}), \bibinfo{pages}{19—24}.
\newblock
\showISSN{0098-9134}
\urldef\tempurl%
\url{https://doi.org/10.3928/00989134-20070701-04}
\showDOI{\tempurl}


\bibitem[\protect\citeauthoryear{Schuster and Paliwal}{Schuster and
  Paliwal}{1997}]%
        {paper23}
\bibfield{author}{\bibinfo{person}{Mike Schuster} {and}
  \bibinfo{person}{Kuldip~K Paliwal}.} \bibinfo{year}{1997}\natexlab{}.
\newblock \showarticletitle{Bidirectional recurrent neural networks}.
\newblock \bibinfo{journal}{\emph{IEEE transactions on Signal Processing}}
  \bibinfo{volume}{45}, \bibinfo{number}{11} (\bibinfo{year}{1997}),
  \bibinfo{pages}{2673--2681}.
\newblock


\bibitem[\protect\citeauthoryear{Vaswani, Shazeer, Parmar, Uszkoreit, Jones,
  Gomez, Kaiser, and Polosukhin}{Vaswani et~al\mbox{.}}{2017}]%
        {paper17}
\bibfield{author}{\bibinfo{person}{Ashish Vaswani}, \bibinfo{person}{Noam
  Shazeer}, \bibinfo{person}{Niki Parmar}, \bibinfo{person}{Jakob Uszkoreit},
  \bibinfo{person}{Llion Jones}, \bibinfo{person}{Aidan~N Gomez},
  \bibinfo{person}{{\L}ukasz Kaiser}, {and} \bibinfo{person}{Illia
  Polosukhin}.} \bibinfo{year}{2017}\natexlab{}.
\newblock \showarticletitle{Attention is all you need}. In
  \bibinfo{booktitle}{\emph{Advances in neural information processing
  systems}}. \bibinfo{pages}{5998--6008}.
\newblock


\bibitem[\protect\citeauthoryear{Wu, Bell, and Wodchis}{Wu
  et~al\mbox{.}}{2012}]%
        {paper3}
\bibfield{author}{\bibinfo{person}{Chen Wu}, \bibinfo{person}{Chaim~M Bell},
  {and} \bibinfo{person}{Walter~P Wodchis}.} \bibinfo{year}{2012}\natexlab{}.
\newblock \showarticletitle{Incidence and economic burden of adverse drug
  reactions among elderly patients in Ontario Emergency Departments}.
\newblock \bibinfo{journal}{\emph{Drug safety}} \bibinfo{volume}{35},
  \bibinfo{number}{9} (\bibinfo{year}{2012}), \bibinfo{pages}{769--781}.
\newblock


\bibitem[\protect\citeauthoryear{Zelenko, Aone, and Richardella}{Zelenko
  et~al\mbox{.}}{2003}]%
        {paper8}
\bibfield{author}{\bibinfo{person}{Dmitry Zelenko}, \bibinfo{person}{Chinatsu
  Aone}, {and} \bibinfo{person}{Anthony Richardella}.}
  \bibinfo{year}{2003}\natexlab{}.
\newblock \showarticletitle{Kernel methods for relation extraction}.
\newblock \bibinfo{journal}{\emph{Journal of machine learning research}}
  \bibinfo{volume}{3}, \bibinfo{number}{Feb} (\bibinfo{year}{2003}),
  \bibinfo{pages}{1083--1106}.
\newblock


\end{thebibliography}

\end{document}